# An Innovative Scheme For Effectual Fingerprint Data Compression Using Bezier Curve Representations

Vani Perumal
Department of Computer Applications
S.A.Engineering College, Chennai – 600 077, India.

Dr.Jagannathan Ramaswamy
Deputy Registrar (Education)
Vinayaka Missions University, Chennai, India.

*Abstract*— Naturally, with the mounting application of biometric systems, there arises a difficulty in storing and handling those acquired biometric data. Fingerprint recognition has been recognized as one of the most mature and established technique among all the biometrics systems. In recent times, with fingerprint recognition receiving increasingly more attention the amount of fingerprints collected has been constantly creating enormous problems in storage and transmission. Henceforth, the compression of fingerprints has emerged as an indispensable step in automated fingerprint recognition systems. Several researchers have presented approaches for fingerprint image compression. In this paper, we propose a novel and efficient scheme for fingerprint image compression. The presented scheme utilizes the Bezier curve representations for effective compression of fingerprint images. Initially, the ridges present in the fingerprint image are extracted along with their co-ordinate values using the approach presented. Subsequently, the control points are determined for all the ridges by visualizing each ridge as a Bezier curve. The control points of all the ridges determined are stored and are used to represent the fingerprint image. When needed, the fingerprint image is reconstructed from the stored control points using Bezier curves. The quality of the reconstructed fingerprint is determined by a formal evaluation. The proposed scheme achieves considerable memory reduction in storing the fingerprint.

I. INTRODUCTION

In recent times, biometrics-based verification has been receiving a lot of attention chiefly, because of the unprecedented proportion of identity fraud ensuing in our society and the increasing emphasis on the emerging automatic personal identification applications [8]. The term "biometrics" is derived from the Greek words - "bio" which means life and "metrics" which means to measure. A more detailed definition of biometrics is "any automatically measurable, robust and unique physical characteristic or personal trait, which can be made use of to recognize an individual or verify the claimed identity of an individual" [1]. The technique of biometric identification is favored over traditional methods involving passwords and PINs (Personal Identification Numbers) for a number of reasons, 1) the person to be verified is to be physically present at the point of identification and/or 2) identification based on biometric techniques alleviates the need to bear in mind a password or carry a token [14].

One good application area of biometrics is forensics; another important application area is criminal identification and prison security. Moreover, it has the impending to be utilized in a large range of civilian application areas. The most frequently used biometric traits include fingerprint, face, iris, hand geometry, voice, palmprint, handwritten signatures and gait [38]. A various other modalities are in different stages of development and assessment [1]. Amongst all the biometric traits, fingerprints possibly possess the highest level of reliability and have been widely used by forensic experts in criminal investigations [15]. It is also used by the police departments around the world to identify suspects and bodies. Because of their uniqueness and immutability, fingerprints, at present are the most widely used biometric features [10]. Other factors that make fingerprint verification one of the most reliable means of biometric authentication is its universality, distinctiveness, permanence and accuracy [6, 7, 24]. Moreover, the price of fingerprint recognition systems has been cost-effective enough to make its way into public use.

Generally, there are two kinds of biometric systems: identification and verification. In identification systems, a biometric signature of a relatively new person is offered to a system. The system matches the new biometric signature with a database of existing biometric signatures of known individuals to identify if the person is previously known or else a stranger. In verification systems, a user offers a biometric signature and the system verifies, if the biometric signature belongs to the claimed identity [2]. Both these systems necessitate the storage of huge number of biometric templates for accomplishing effective recognition of identity. With the rising usage of fingerprint recognition systems the inquiry arises naturally how to store and manage the obtained fingerprint data [18]. Generally, the databases of fingerprint recognition systems may contain millions of fingerprint images. Moreover, a fingerprint image itself consists of massive amounts of data; the storage of fingerprint image databases necessitates huge secondary storage devices [39]. So as to lessen the escalating demand on storage space, efficient data compression techniques are badly desirable [11], [39]. Recently, the compression of fingerprints has gained enormous popularity in automated fingerprint recognition systems due to the increasing number of the fingerprint records in their databases.

There are numerous image compression techniques existing for image compression, like DCT, JPEG, Sub-band Coding, JPEG2000, Wavelet and more [13]. The general objective of all these techniques is to accomplish high compression ratio. In spite of some really good compression algorithms, there is still a need to develop more efficient algorithms for fingerprint images [12]. The chief difficulty in developing compression algorithms for fingerprint is the necessity to achieve minutiae preservation i.e. ridges endings and bifurcations, which are afterwards used in identifications [17]. Literature presents with several different methods for fingerprint compression. They can be categorized in two divisions: 1) Fingerprint data compression techniques based on extraction and compression of essential details in fingerprints like ridges and/or features. Some studies in this category are Abdelmalek et al. [40], Chong et al. [41], Yamada et al. [42] and Costello et al. [43]. 2) Fingerprint image compression techniques based on image transformations that are tuned for fingerprint images. A small number of them also use several vector quantization techniques [44]. The techniques in this category cannot exploit the regular structural properties of fingerprints to accomplish higher compression ratios.

In the second category, Hopper et al. [45], Bradley et al. [46] and Brislawn et al. [11] studied wavelet/scalar quantization which is used as a standard algorithm by FBI [9]. Numerous other compression methods utilizing wavelets have been reported by Kasei et al. [19] and Sherlock et al. [33, 34]. This research is aimed at devising an efficient fingerprint data compression scheme that will extract and compress the essential data (ridges) in fingerprints. Recently, Yuan Huaqiang et al [36] have presented a fingerprint feature extraction algorithm based on curvature of Bezier curves. Primarily in their algorithm, the ridges in the fingerprint images were traced and then those ridges are fit with Bezier curves. The proposed fingerprint data compression scheme drives motivation from the work of Yuan Huaqiang et al [36].

This paper describes a novel and efficient compression scheme for fingerprint images using Bezier representations. The proposed compression scheme is designed in a way to preserve the fine details in the fingerprint images such as ridge endings and bifurcations. The Bezier curve representations are employed in the presented scheme for achieving better compression with some cost to quality. Initially, the ridges are extracted from the fingerprint image along with their co-ordinate values using the approach discussed. The 'regionprops', a function to measure properties of regions, in MATLAB's Image Processing Toolbox is utilized in the extraction of the ridges. Subsequently, control points of all the ridges are determined by visualizing each ridge as a Bezier curve. The control points consist of a starting point, ending point and two selected co-ordinate values. Afterwards, the control points determined for all the ridges are stored in a file to represent the fingerprint image. These control points are to be utilized in the reconstruction of fingerprint image using Bezier curves. The presented scheme considerably reduces the memory needed for storing a fingerprint biometric template, from KiloBytes (KB) to bytes. The quality of the reconstructed fingerprint is determined using a formal evaluation. The experimental results demonstrate the effectiveness of the proposed scheme in compressing fingerprint images with a better compression ratio and reasonable reconstruction quality.

The rest of the paper is organized as follows: A brief review of the recent researches related to fingerprint image compression is given in Section II. An introduction to Bezier curves is provided in Section III. The novel and efficient compression scheme proposed for compressing fingerprint images is discussed along with the extraction of ridges in Section IV. The reconstruction procedure of the Bezier curves is discussed in section V. The experimental results are presented in Section VI. Finally, the conclusions are summed up in Section VII.

## II. REVIEW OF RELATED RESEARCHES

A handful of researchers have presented approaches for the compression of fingerprint images. With fingerprint image databases growing at rapid rate, developing schemes for the compression of fingerprint images has emerged as an active and eminent research area. A brief review of some recent and significant researches is presented here.

Awad Kh. Al-Asmari [16] has implemented a progressive fingerprint image compression method (for storage or transmission) by means of edge detection scheme. The image was decomposed into two components, the first component is called as the primary component which encloses the edges and the second component contains the textures and the features. An approximate of the image was reconstructed in the first stage at a bit rate of 0.0223 bpp for one Sample and 0.0245 bpp for another Sample image. The quality of the reconstructed images was competitive to the 0.75 bpp target bit set by FBI standard. The compression ratio for the algorithm is about 45:1 (0.180 bpp).

S. S. Gornale *et al.* [17] have highlighted different transforms of wavelet packet and their compression ratio for noisy and noiseless fingerprint images. They have also showed that the compression ratio can be increased by selecting appropriate threshold value. The compression ratios of noisy and noiseless fingerprint images are found by considering the number of zeros and the retain energy. Wavelet packet transform certainly has an effect on the Retain Energy (RE) and Number of Zeros (NZ) but the extent of it is dependent on the decomposition level, the type of image, threshold and also the type of transform used. For a maximum threshold value and greater level of decomposition, more energy can be lost, since, at higher levels of decomposition there is a higher proportion of the coefficients in the detail sub-signals. Therefore, it is always crucial to choose an optimal threshold value so as to achieve better compression and minimum loss to images.

S. Esakkirajan *et al.* [21] have presented an approach based contourlet transform and multistage vector quantization for the compression of fingerprint images. An extensive result has been taken on different types of fingerprints. It can be seen that the PSNR (peak signal to noise ratio) obtained by contourlet transform was higher than that of wavelet transform. Therefore, a better image reconstruction was achievable with less number of bits, by using contourlet transform. The experimental results proved the fact that MSVQ (multistage vector quantization) was appropriate for low bit rate image coding. The proposed system yields encoding outputs of good quality around 0.5 bits per dimension (bpd) and very good results at around 1 bpd. One possible and easy extension to the proposed scheme is to include more stages in MSVQ in order to increase the output image quality.

R. Sudhakar *et al.* [22] have focused a great deal on the compression scheme by integrating wavelet Footprints with Non-linear Approximation (NLA). The results of the Compression Algorithm based on wavelet footprints portrayed that there was a gradual progression in the Compression Ratio (CR) and Peak Signal to Noise Ratio (PSNR) over the set partitioning in Hierarchical Trees (SPIHT) Algorithm. Both the results, theoretical and experimental, have proved the potential of the scheme.S.S.Gornale *et al.* [23] has evaluated to identify the best of the bi-orthogonal wavelet filter from Daubechies, Symlet and Coiflet for lossy fingerprint image compression and they have used it through different orders at 1 to 5 decomposition levels on the fingerprint images. The results have shown that the Coiflet4 (4th order) wavelet filter is more appropriate for lossy fingerprint image compression and provides an enhanced compression at 5th level.

Gulzar A. Khuwaja [35] has identified the best design parameters for a data compression scheme designed for fingerprint images. Their method focuses on reducing the transmission cost while maintaining the person's identity. In choosing the wavelet packet's filters, decomposition level, and sub-bands that are better adapted to the frequency characteristics of the image, one may achieve better image representation in the sense of lower entropy or minimum distortion is considered. Empirical results proved that the selection of the best parameters has a remarkable effect on the data compression rate of fingerprint images. Statistical significance test was conducted on the experimental measures to perform the most suitable wavelet shape for fingerprint images. Image quality measures such as mean square error and peak signal-to-noise ratio are used to estimate the performance of different wavelet filters.

Song Zhao and Xiao-Fei Wang [20] have presented a compression algorithm termed as, Wavelet-Based Contourlet Transform (WBCT), for fingerprint images. It is based on wavelet transform and directional filter banks (DFBs) and can be used for efficiently approximating natural images containing contours and oscillatory patterns. To minimize frequency scrambling, a scheme based on maximally-flat filters which implements the DFBs was proposed. A quadtree sorting procedure, similar to SPIHT, is used to explicitly form classes of WBCT coefficients. The classes are encoded using arithmetic and trellis-coded quantization. The resulting encoding algorithm presents constant improvement over SPIHT performance. Simulations reveal that the new encoding algorithm gives enhanced encoding performance over SPIHT and preserves more fingerprint image details.

Kasaei. S *et al.* [11] have presented a vector quantization scheme based on an accurate model for the distribution of the wavelet coefficients and a compression algorithm for fingerprint images using wavelet packets and lattice vector quantization. This technique is based on the generalized Gaussian distribution. They also discussed a method for determining the largest radius of the lattice used and its scaling factor, for both uniform and piecewise-uniform pyramidal lattices. The presented algorithm aims to achieve the best rate-distortion function by adapting to the characteristics of the sub-images. In the optimization algorithm, no assumptions about the lattice parameters are made, and no training and multi-quantizing are required. They proved that the wedge region problem encountered with sharply distributed random sources was resolved in the proposed algorithm. The proposed algorithm adjusts to variability in input images and to the specified bit rates. Compared to other available image compression algorithms, the proposed algorithm results in high quality reconstructed images for identical bit rates.

Fingerprint feature extraction is the main step of fingerprint identification. Yuan Huaqiang *et al.* [36] have proposed a feature extraction algorithm, which describes the fingerprint features with the bending information of fingerprint ridges. Firstly, the ridges in the specific region of fingerprint images are traced by the algorithm, and then, these ridges are fit with Bezier curve. Finally, the point that has the maximal curvature on Bezier curve was defined as a feature point. Experimental results confirmed that these kinds of feature points characterize the bending trend of fingerprint ridges efficiently, and they are robust to noise. Also, the extraction accuracy of the algorithm is superior to the conventional approaches.

III. BEZIER CURVES

Originally, Bezier curves were introduced in 1959 by Paul de Casteljau. But, only in the 1970's, when Pierre Bezier, French engineer at Renault, utilized them to design automobiles, they emerged as a famous shape. Presently, Bezier curves are extensively utilized in many fields: industrial and computer-aided design, vector-based drawing, font design (especially in PostScript font) and 3D modeling [29]. Bezier curves are also being made use of in computer graphics to model smooth curves. Since the curve is entirely contained in the convex hull of its control points, it is possible to graphically display the points and also the control points can be used to manipulate the curve intuitively. A Bezier curve lets you to state, not only the end points of the line, but also the course of the line as it goes by the end points. Bezier curves of the third order are the most commonly used and can be completely defined by four points: two endpoints ($P_1$, $P_4$) and two control points ($P_2$, $P_3$). The control points are not positioned on the curve itself but they define the shape of the curve [25]. Considering Figure 1, the Bezier curve defined starts at $P_1$, goes toward $P_2$ and arrives at $P_4$ coming from the direction of $P_3$. Generally, these Bezier curves do not pass through the control points $P_2$ or $P_3$. Such a curve is called cubic Bezier curve.

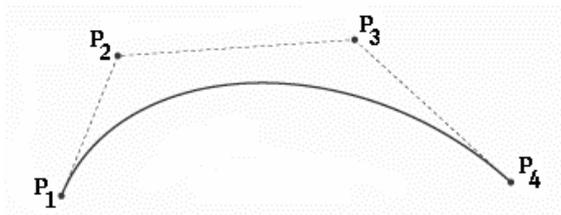

Figure 1. Cubic Bezier Curve

Clearly, a cubic Bezier curve is a function of four points, of which two will be the end points of the curve and the other two will be the points lying outside the curve. These set of four points specify how the entire curve can be built in entirety. The reverse method is employed for regenerating the curve from the control points. In this research, so as to reduce the memory overhead incurred in fingerprint storage, each ridge in the fingerprint can be visualized as a cubic Bezier curve and can be stored in as four points (two end points and two control points).

IV. NOVEL AND EFFICIENT COMPRESSION SCHEME FOR FINGERPRINT IMAGES

The proposed novel scheme for effective compression of fingerprint images is described in this section. The proposed scheme consists of two steps:

1) Extraction of ridges along with their co-ordinate values and

2) Compression using Bezier curve representations.

The major steps involved in the proposed fingerprint image compression scheme are shown in Figure 2.

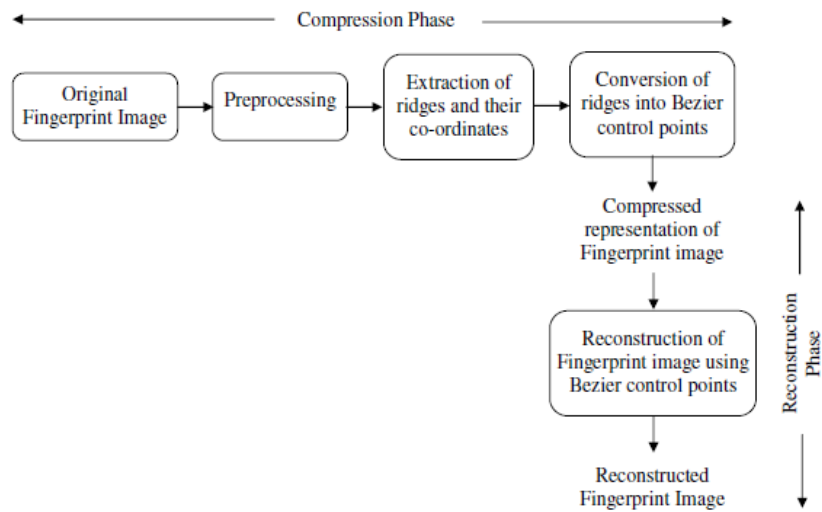

Figure 2. Block diagram of the Proposed Compression Scheme

*A. Extraction of Ridges with Their Co-Ordinates*

The extraction of ridges and their co-ordinate values from the fingerprint image is discussed in this sub-section. A fingerprint can be defined as a pattern of friction ridges on the surface of a fingertip. Minutiae are local discontinuities in the fingerprint pattern that symbolize terminations and bifurcations. The point where a ridge ends abruptly is called the ridge termination and the point where a ridge forks or diverges into branch ridges is called the ridge bifurcation [3]. The ridge structures in fingerprint images are not at all times well defined, and hence, an enhancement algorithm that can enhance the clarity of the ridge structures, is essential [32]. Every ridge or a portion of a line in a fingerprint is classified into one of the three different major patterns; a loop or a whorl or an arch. A loop pattern in a fingerprint can be shown when the ridges start on one side of the finger, reach the center of the finger (core point) and then "loop" back to the same side. A whorl pattern can be identified as the concentric circles that are formed by the ridges in the center of one's finger. The remainder of these ridges shape themselves around this whorl pattern. Finally, the arch pattern: where the ridges start at one side of the finger and span themselves across the center of the finger to the other side [4, 5]. The major steps involved in the extraction of the ridges and their co-ordinate values are,

- Preprocessing
- Ridge Extraction

*1) Preprocessing:* The preprocessing steps involved in the extraction of the ridges from the fingerprint image are namely:
- Histogram Equalization
- Fast Fourier Transform (FFT) Enhancement
- Binarization
- Orientation Field Estimation
- Region of Interest (ROI) Extraction by Morphological operations

**(i) Histogram Equalization**

Histogram equalization describes a mapping of grey levels p into grey levels q in such a way that the distribution of grey level q is uniform. This mapping stretches contrast (expands the range of grey levels) for grey levels near the histogram maxima. As the contrast is extended to most of the image pixels, the transformation increases the detectability of many image features [26]. A fingerprint image and its histogram equalized output are shown in Figure 3.

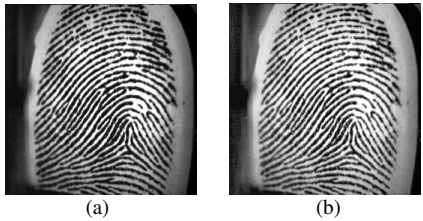

(a)          (b)

Figure 3. (a) Original Fingerprint (b) Histogram equalized Fingerprint

**(ii) Fast Fourier Transform (FFT) Enhancement**

The image enhancement techniques are frequently used to decrease the noise and improve the definition of ridges against valleys. In our scheme, to enhance the fingerprint image, Fast Fourier Transform (FFT) is applied separately to each block of the image [26]. The enhanced image is then binarized and fed as input to orientation field estimation.

**(iii) Binarization**

Binarization increases the contrast between the ridges and valleys in a fingerprint image, and as a result eases minutiae extraction. The binarization process involves,

a) Investigating the grey-level value of each pixel in the enhanced fingerprint image, and

b) If the value is greater than the global threshold, then the pixel value is set to a binary value one; otherwise, it is set to zero.

The binarization results a binary fingerprint image containing two levels of information, the foreground ridges and the background valleys [27].

**(iv) Orientation Field Estimation**

The orientation field of a fingerprint image defines the local orientation of the ridges contained in that the fingerprint. The orientation estimation is an elementary step in the enhancement process as the succeeding filtering stage depends on the local orientation so as to efficiently enhance the fingerprint image [27]. Principally, there exists two methodologies to compute the orientation field of fingerprint namely, 1) filter-bank based approaches and 2) gradient-based approaches. The proposed scheme for fingerprint compression employ gradient based approach for estimating the orientation field of the fingerprint image. Primarily, the gradient vectors are computed by considering the partial derivatives of image intensity at every pixel. The gradient vectors can be represented as $[g_x, g_y]^T$. With an input fingerprint image, the gradient vectors signify the highest deviation of gray intensity that lie perpendicular to the edge of ridge lines [31].

**(v) Morphological Operations**

The binary morphological operators are applied on the binarized fingerprint image. The primary function of the morphological operators is the elimination of any obstacles and noise from the image. In addition, the morphological operators remove the unnecessary spurs, bridges and line breaks. Then, thinning process is performed to reduce the thickness of the lines (removes redundant pixels) so that the lines become 1-pixel wide and easily distinguishable from the other regions of the image. Clean operator, Hbreak operator, Spur operator and Thinning are the morphological operators utilized in the proposed scheme [30], [28]. The result of morphological operations is depicted in Figure 4.

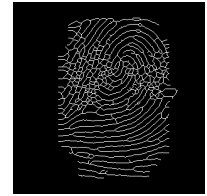

Figure 4. Fingerprint Image after Morphological Operations

*2) Extraction of Ridges:* The preprocessing steps result with a fingerprint image that gives a clear depiction of foreground ridges. Subsequently, the ridges present in the preprocessed fingerprint image are separated and their co-ordinate values are found using the following steps:

1. The preprocessed fingerprint image is likely to have some connected ridges that might affect the extraction of ridges. Hence, we first segregate those connected ridges based on the minutiae points. Here, the ridge thinning algorithm is utilized for minutiae points' extraction [38]. In the ridge thinning algorithm, the image is first divided into two dissimilar subfields that show a likeness to a checkerboard pattern. In the initial sub iteration, only when all three conditions, G1, G2, and G3 are satisfied the pixel p from the initial subfield is removed. Whereas, in the second sub iteration, only when all three conditions, G1, G2, and G3' are satisfied, the pixel p from the foremost subfield is removed.

**Condition G1:**

$$X_H(P) = 1$$

Where

$$X_H(P) = \sum_{i=1}^{4} b_i$$

$$b_i = \begin{cases} 1 \text{ if } x_{2i-1} = 0 \text{ and } (x_{2i} = 1 \text{ or } x_{2i+1} = 1) \\ 0 \text{ otherwise} \end{cases}$$

$x_1, x_2, ..., x_8$ are the values of the eight neighbors of $p$, starting with the east neighbor and numbered in counter-clockwise order.

**Condition G2:**

$$2 \leq \min\{n_1(p), n_2(p)\} \leq 3$$

where

$$n_1(p) = \sum_{k=1}^{4} x_{2k-1} \vee x_{2k}$$

$$n_2(p) = \sum_{k=1}^{4} x_{2k} \vee x_{2k+1}$$

**Condition G3:**
$$(x_2 \vee x_3 \vee \overline{x}_8) \wedge x_1 = 0$$

**Condition G3':**
$$(x_6 \vee x_7 \vee \overline{x}) \wedge x_5 = 0$$

One iteration of the thinning algorithm combines the two subiterations.

2. The pixel locations corresponding to the minutiae points are replaced with black pixels. The resultant fingerprint image contains the ridges; each separated from the other.

3. The fingerprint image obtained in Step 2 is fed as input to the regionprops function of MATLAB's Image Processing Toolbox [47]. The regionprops function determines the properties of each of the individual ridges $R$ present in the fingerprint image $I$.

$$R[i] = regionprops(I); i \rightarrow no\ of\ ridges$$

4. The individual ridges are extracted and the properties of the individual ridges are utilized to acquire the co-ordinate values $C_v$ of the ridges.

*B. Compression Using Bezier Curve Representations*

A fingerprint image can have hundreds of ridges each having its own structure. In the proposed scheme, each ridge is visualized as a cubic Bezier curve and its Bezier control points (two end points and two control points) are determined. The set of four Bezier control points determined, serve as compressed form of an individual ridge. So, every fingerprint image with *n* ridges can be compressed into a file containing *4*n* Bezier control points. The different structures of the ridges present in the original fingerprint image are shown in Figure 5.

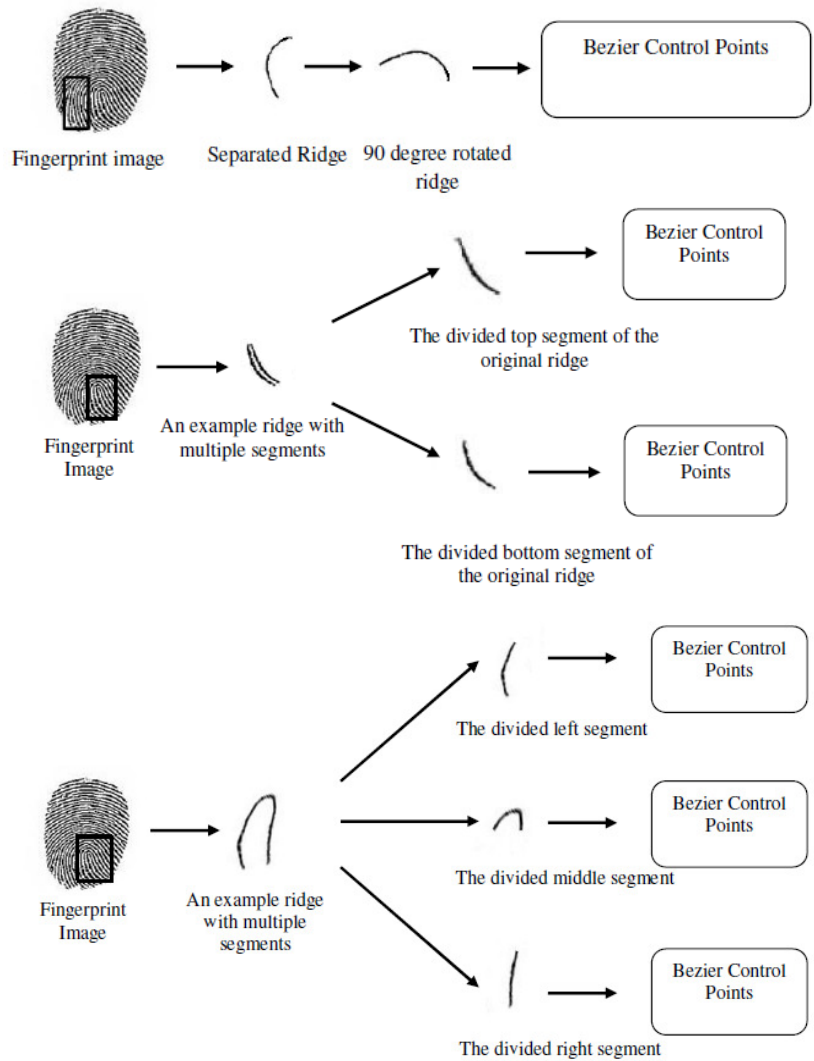

Figure 5. Different structures of the ridges

Subsequently, for every ridge visualized as a Bezier curve,

1) Determine two end points namely, $P_1$ (the origin or starting point) and $P_2$ (the destination or terminating point), from $C_v$.

2) Compute two control points that directs the construction of the Bezier curve. The two control points are, $P_2$ (control point towards which the Bezier curve moves from $P_1$) and $P_3$ (control point that directs the Bezier curve towards $P_4$).

The Bezier control points of the ridges are determined by bearing in mind the following factors: The Bezier curve at all times passes through the end points and lies within the convex hull of the control points. The curve is tangent to $P_2$-$P_1$ and $P_n$-$P_{n-1}$ the endpoints. The "variation diminishing property" of these curves is that no line can have more intersections with a Bezier curve than with the curve obtained by joining consecutive points with straight line segments. A desirable property of these curves is that the curve can be translated and rotated by performing these operations on the control points.

It is sufficient to store all the four Bezier control points instead of storing the actual Bezier curve i.e. ridge of the fingerprint. Also, the original ridge in the fingerprint can be reproduced from these stored control points by the properties of the Bezier curve. Thus, the proposed scheme for fingerprint compression achieves an effective reduction in the memory space required to store the fingerprint.

V. RECONSTRUCTION OF FINGERPRINT IMAGE FROM BEZIER CONTROL POINTS

The encoded fingerprints comprise of a set of control points, each corresponding to individual ridges of the original fingerprint. For reconstruction or decoding of the compressed fingerprint, every individual ridge in the fingerprint is decoded using the properties of the Bezier curve. All these decoded ridges finally unite to form the fingerprint image. The input to the reconstruction is a set of control points from which the Bezier curve is to be constructed. The mathematical formulation of the Bezier curve construction from the control points is as follows [37]:

Given $N+1$ control points $P_k$ with $k = 0\ to\ N$. The Bezier parametric curve is given by $B(u)$,

$$B(u) = \sum_{k=0}^{N} P_k \frac{N!}{k!(N-k)!} u^k (1-u)^{N-k} \quad 0 \leq u \leq 1$$

Since, Bezier curves are parametric curves the above formula is applied independently to the $x$ and $y$ coordinates of a point in a 2D curve [37].

$$\left. \begin{aligned} B(u).x &= \sum_{k=0}^{N} (P_k.x) \frac{N!}{k!(N-k)!} u^k (1-u)^{N-k} \\ B(u).y &= \sum_{k=0}^{N} (P_k.y) \frac{N!}{k!(N-k)!} u^k (1-u)^{N-k} \end{aligned} \right\} \quad 0 \leq u \leq 1$$

In our scheme, we employ cubic Bezier curves, which can very well be illustrated by four control points. So, given 4 controls points $(p_0, p_1, p_2, p_3)$, we derive the mathematical formula of the cubic Bezier as follows [37]:

$$B(u) = \sum_{k=0}^{3} P_k \frac{3!}{k!(3-k)!} u^k (1-u)^{3-k}$$

$$B(u) = P_0(1-u)^3 + 3P_1 u(1-u)^2 + 3P_2 u^2(1-u) + P_3 u^3$$

$$B(u) = u^3(-P_0 + 3P_1 - 3P_2 + P_3) + u^2(3P_0 - 6P_1 + 3P_2) + u(-3P_0 + 3P_1) + P_0$$

$$B(u) = u^3(P_3 + 3(P_1 - P_2) - P_0) + 3u^2(P_0 - 2P_1 + P_2) + 3u(P_1 - P_0) + P_0$$

VI. EXPERIMENTAL RESULTS

The experimental results of the novel and efficient scheme presented for compressing fingerprint images are provided in this section. The proposed scheme is implemented using Matlab (Matlab 7.4) and Java. The ridges present in the fingerprint image are first extracted by using Matlab. Then, the process of determining the Bezier control points and the reconstruction phase are performed using Java. First, the ridges in the preprocessed fingerprint image are separated with their respective co-ordinate values. Subsequently, each ridge is visualized as a Bezier curve and for every curve, four control points are determined. The set of four control points represent the compressed form of an individual ridge. Consequently, using the Bezier control points, we have reconstructed the fingerprint image, which preserves the fine details of the original fingerprint image. The results obtained from experimentation with two fingerprint images are shown in Figure 6 and 7. Each figure consists of a) the original fingerprint image, b) image constructed from the co-ordinate values of the extracted ridges, c) the reconstructed image using Bezier control points and d) Evaluation result (image (b) superimposed on image (c)). The performance of the presented scheme has been evaluated by superimposing the fingerprint image constructed using the co-ordinate values of the extracted ridges on the reconstructed fingerprint image using Bezier control points

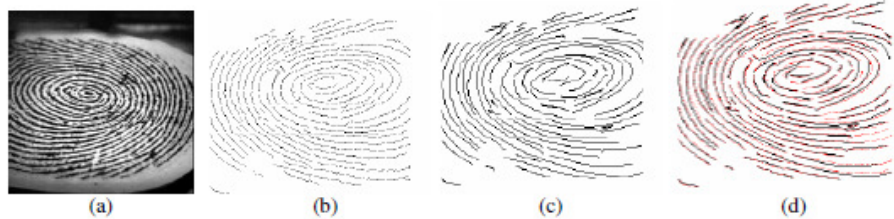

Figure 6. (a) Original Fingerprint Image 1 (b) Constructed Fingerprint image using co-ordinates of the ridges (c) Reconstructed fingerprint image using Bezier control points (d) Evaluation Result

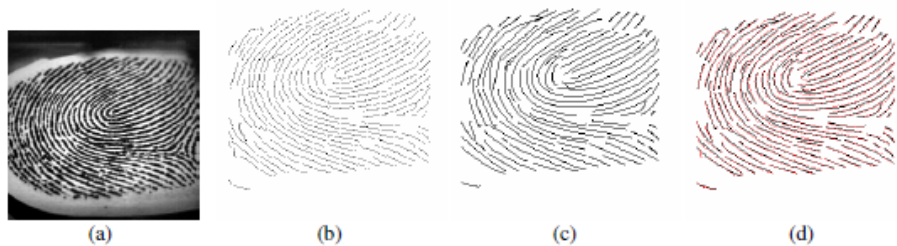

Figure 7. (a) Original Fingerprint Image 2 (b) Constructed Fingerprint image using co-ordinates of the ridges (c) Reconstructed fingerprint image using Bezier control points (d) Evaluation Result

Moreover, the proposed fingerprint compression scheme has achieved an exceptionally good compression ratio. The results obtained in compressing two fingerprint images are given in Table 1.

TABLE I: COMPRESSION RESULTS OF THE PROPOSED SCHEME

| Image | Original image size (KB) | Compressed file size (KB) |
|---|---|---|
| Fingerprint 1 | 19.4 | 2.7 |
| Fingerprint 2 | 19.2 | 2.6 |

## VII. CONCLUSION

In this paper, we have presented a novel and efficient compression scheme using Bezier curves for compressing fingerprint images. Initially, the ridges and their co-ordinate values in the fingerprint image have been extracted with the aid of the approach discussed in the paper. Subsequently, the control points for all the extracted ridges have been determined by visualizing the ridges as Bezier curves. The determined control points have been stored in a file that represents the compressed counterpart of the fingerprint image. The fingerprint images are reconstructed, when needed, from the stored control points using Bezier curves. The presented scheme has achieved better compression with some cost to accuracy. The experimental results have demonstrated the effectiveness of the presented compression scheme.

AUTHORS PROFILE

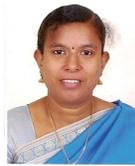

**Vani Perumal** received the B.Sc degree from the Department of Computer Science, University of Madras, the M.C.A degree from the Department of Computer Applications, Bharathidasan University, the M.Phil degree from the Department of Computer Science, Mother Teresa Women's University. She is currently pursuing Ph.D degree in Computer Science, specialization in Image Processing and Pattern Recognisation at Mother Teresa Women's University, Tamil Nadu, India. From 2002 to 2006, she was the Head in charge of Computer Science Department, Soka Ikeda College, Chennai. She is currently working with S.A.Engineering College, Chennai, India.

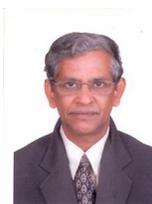

**Jagannathan Ramaswamy** received B.Sc, M.Sc and Ph.D degrees from the University of Madras, India He obtained his Master of Philosophy degree in Space Physics from Anna University, Chennai. He was the Reader and the Head of the Postgraduate Department of Physics at D.G.Vaishnav College, Chennai. Dr.Jagannathan is currently the Chairman cum Secretary of India Society of Engineers, Madras Chapter, Managing Editor (Publications), Asian Journal of Physics and Deputy Registrar (Education), Vinayaka Missions University, Chennai, India.